\documentclass[conference]{IEEEtran}
\IEEEoverridecommandlockouts

\ifCLASSINFOpdf
\else
\fi

\usepackage[dvips]{graphicx}
\usepackage{indent}
\usepackage{amsmath,amssymb}
\usepackage{paralist}
\usepackage{eclbkbox}
\usepackage{subfigure}
\usepackage{url}

\makeatletter
\def\tbcaption{\def\@captype{table}\caption}
\def\figcaption{\def\@captype{figure}\caption}

\newcommand{\bvec}[1]{\mbox{\boldmath $#1$}}

\begin{document}
\title{An Adaptive Learning Method of Deep Belief Network by Layer Generation Algorithm
\thanks{\copyright 2016 IEEE. Personal use of this material is permitted. Permission from IEEE must be obtained for all other uses, in any current or future media, including reprinting/republishing this material for advertising or promotional purposes, creating new collective works, for resale or redistribution to servers or lists, or reuse of any copyrighted component of this work in other works.}}

\author{
\IEEEauthorblockN{Shin Kamada}
\IEEEauthorblockA{
Graduate School of Information Sciences, \\
Hiroshima City University\\
3-4-1, Ozuka-Higashi, Asa-Minami-ku,\\
Hiroshima, 731-3194, Japan\\
Email: da65002@e.hiroshima-cu.ac.jp}
\and
\IEEEauthorblockN{Takumi Ichimura}
\IEEEauthorblockA{Faculty of Management and Information Systems,\\
Prefectural University of Hiroshima\\
1-1-71, Ujina-Higashi, Minami-ku,\\
Hiroshima, 734-8559, Japan\\
Email: ichimura@pu-hiroshima.ac.jp}
}

\maketitle

\begin{abstract}
Deep Belief Network (DBN) has a deep architecture that represents multiple features of input patterns hierarchically with the pre-trained Restricted Boltzmann Machines (RBM). A traditional RBM or DBN model cannot change its network structure during the learning phase. Our proposed adaptive learning method can discover the optimal number of hidden neurons and weights and/or layers according to the input space. The model is an important method to take account of the computational cost and the model stability. The regularities to hold the sparse structure of network is considerable problem, since the extraction of explicit knowledge from the trained network should be required. In our previous research, we have developed the hybrid method of adaptive structural learning method of RBM and Learning Forgetting method to the trained RBM. In this paper, we propose the adaptive learning method of DBN that can determine the optimal number of layers during the learning. We evaluated our proposed model on some benchmark data sets.
\end{abstract}

\IEEEpeerreviewmaketitle

\section{Introduction}
\label{sec:Introduction}
Deep Learning representation attracts many advances in a main stream of machine learning in the field of artificial intelligence (AI) \cite{Bengio09}. Especially, the industrial circles are deeply impressed with the advanced outcome in　AI techniques to the increasing classification capability of Deep Learning. That is a misunderstanding that a peculiarity is the multi-layered network structure. The impact factor in Deep Learning enables the pre-training of the detailed features in each layer of hierarchical network structure. The pre-training means that each layer in the hierarchical network architecture trains the multi-level features of input patterns and accumulates them as the prior knowledge. Restricted Boltzmann Machine (RBM) \cite{Hinton12} is an unsupervised learning method that realizes high capability of representing a probability distribution of input data set. Deep Belief Network (DBN) \cite{Hinton06} is a hierarchical learning method by building up the trained RBMs.

We have proposed the theoretical model of adaptive learning method in RBM \cite{Kamada16_ICONIP}. Although a traditional RBM model cannot change its network structure, our proposed model can discover the optimal number of hidden neurons and their weights during training RBM by the neuron generation and the neuron annihilation algorithm. The method monitors the convergence situation with the variance of 2 or more parameters of RBM. Moreover, we developed the Structural Learning Method with Forgetting (SLF) of RBM to hold the sparse structure of network. The method may help to discover the explicit knowledge from the structure and weights of the trained network. The combination method of adaptive learning method of RBM and Forgetting method can show higher classification capability on CIFAR-10 and CIFAR-100 \cite{CIFAR10} in comparison to the traditional model.

However some similar patterns were not classified correctly due to the lack of the classification capability of only one RBM. In order to perform more accuracy to input patterns even if the pattern has noise, we employ the hierarchical adaptive learning method of DBN that can determine the optimal number of hidden layers according to the given data set. DBN has the suitable network structure that can show a higher classification capability to multiple features of input patterns hierarchically with the pre-trained RBM. It is natural extension to develop the adaptive learning method of DBN with self-organizing mechanism.

In \cite{Kamada16_YRW}, the visualization tool between the connections of hidden neurons and their weights in the trained DBN was developed and was investigated the signal flow of feed forward calculation. In this paper, some experimentation of CIFAR-10 with various parameters of RBM and DBN were examined to verify the effectiveness of our proposed adaptive learning of DBN. As a result, the score of classification capability was the top score of world records in \cite{record}.

\section{Adaptive Learning Method of RBM}
\label{sec:AdaptiveRBM}
\subsection{Overview of RBM}
A RBM consists of 2 kinds of neurons: visible neuron and hidden neuron. A visible neuron is assigned to the input patterns. The hidden neurons can represent the features of given data space. The visible neurons and the hidden neurons take a binary signal $\{0,1\}$ and the connections between the visible layer and the hidden layer are weighted the parameters $\bvec{W}$. There are no connections among the hidden neurons, the hidden layer forms an independent probability for the linear separable distribution of data space. The RBM trains the weights and some parameters for visible and hidden neurons to reach the small value of energy function.

Let $v_i (0 \leq i \leq I)$ be a binary variable of visible neuron. $I$ is the number of visible neurons. Let $h_j (0 \leq j \leq J)$ be a binary variable of hidden neurons. $J$ is the number of hidden neurons. The energy function $E(\bvec{v}, \bvec{h})$ for visible vector $\bvec{v} \in \{ 0, 1 \}^{I}$ and hidden vector $\bvec{h} \in \{ 0, 1 \}^{J}$ in Eq.(\ref{eq:energy}). $p(\bvec{v}, \bvec{h})$ in Eq.(\ref{eq:prob}) shows the joint probability distribution of $\bvec{v}$ and $\bvec{h}$.

\begin{equation}
E(\bvec{v}, \bvec{h}) = - \sum_{i} b_i v_i - \sum_j c_j h_j - \sum_{i} \sum_{j} v_i W_{ij} h_j ,
\label{eq:energy}
\end{equation}
\begin{equation}
p(\bvec{v}, \bvec{h})=\frac{1}{Z} \exp(-E(\bvec{v}, \bvec{h})), \;  Z = \sum_{\bvec{v}} \sum_{\bvec{h}} \exp(-E(\bvec{v}, \bvec{h})) ,
\label{eq:prob}
\end{equation}
where the parameters $b_i$ and $c_j$ are the coefficients for $v_i$ and $h_j$, respectively. Weight $W_{ij}$ is the parameter between $v_i$ and $h_j$. $Z$ in Eq.(\ref{eq:prob}) is the partition function that $p(\bvec{v}, \bvec{h})$ is a probability function and calculates to sum over all possible pairs of visible and hidden vectors as the denominator of $p(\bvec{v}, \bvec{h})$. The traditional RBM learns the parameters $\bvec{\theta}=\{\bvec{b}, \bvec{c}, \bvec{W} \}$ according to the distribution of given input data. The Contrastive Divergence (CD) \cite{Hinton02} has been proposed to make a good performance even in a few sampling steps as a faster algorithm of Gibbs sampling method which is satisfied with Markov chain Monte Carlo methods. CD method must be converged if the variance for 3 kinds of parameters $\bvec{\theta}=\{\bvec{b}, \bvec{c}, \bvec{W} \}$ falls into a certain range under the Lipschitz continuous \cite{Carlson15}.

The variance of gradients for these parameters become larger if the learning situation fluctuates from some experimental results\cite{Kamada16a}. However, the $\bvec{b}$ is influenced largely by the distributions of input patterns. Then the 2 kinds of parameters $\bvec{c}$ and $\bvec{W}$ are monitored to avoid the repercussion of uniformity of impact data \cite{Kamada16a}.

\subsection{Neuron Generation and Neuron Annihilation Algorithm}
\label{subsec:AdaptiveRBM}
The adaptive learning method of RBM \cite{Kamada16a} can be self-organized structure to discover the optimal number of hidden neurons and weights according to the features of a given input data set in learning phase. The basic idea of the neuron generation and neuron annihilation algorithm works in multi-layered neural network \cite{Ichimura04} and we introduce the concept into RBM training method. The method measures the criterion of network stability with the fluctuation of weight vector.

\begin{figure}[tbp]
\begin{center}
\subfigure[Neuron Generation]{\includegraphics[scale=0.5]{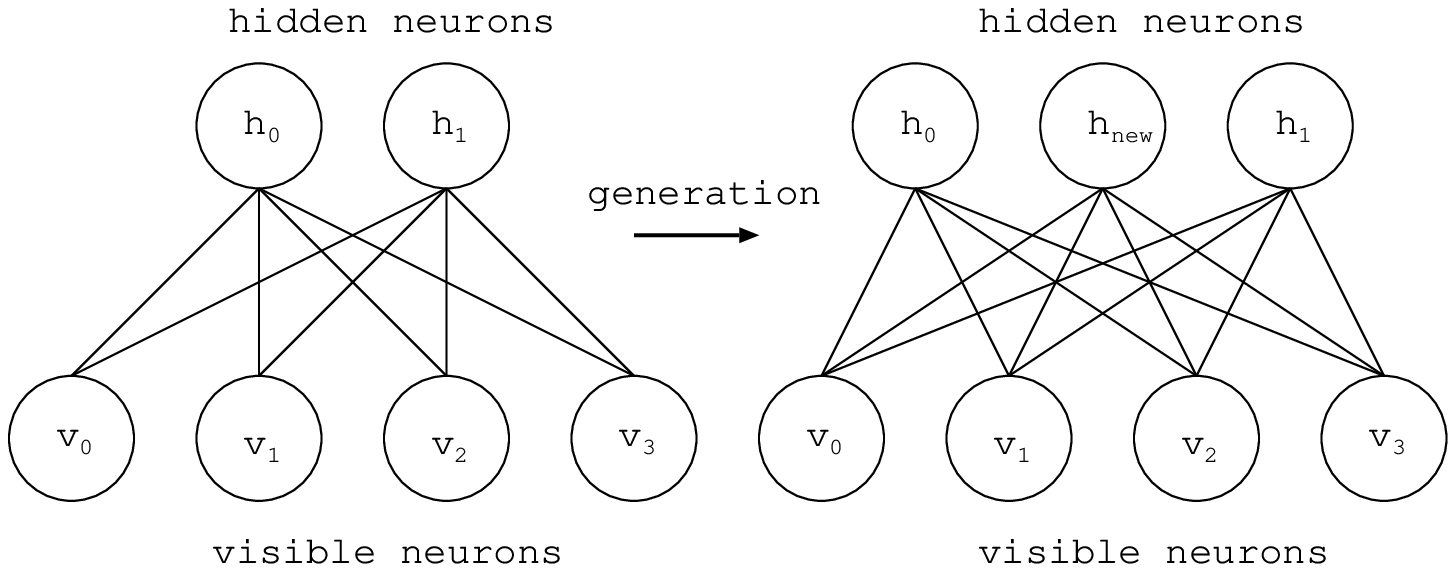}\label{fig:neuron_generation}}
\subfigure[Neuron Annihilation]{\includegraphics[scale=0.5]{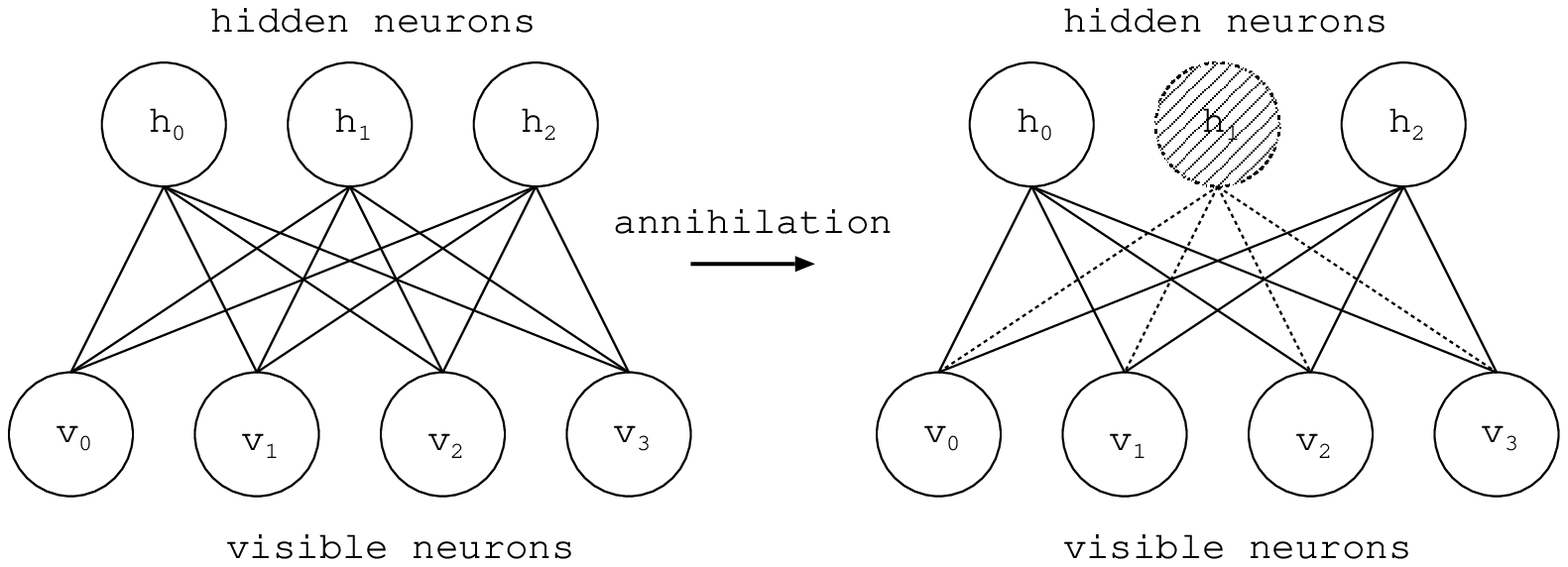}\label{fig:neuron_annihilation}}
\caption{An Adaptive RBM}
\label{fig:adaptive_rbm}
\end{center}
\end{figure}

If there is not enough neurons in a RBM to classify the input patterns, then the weight vector and the other parameters will fluctuate large even after long iterations. We must consider the case that some hidden neurons cannot afford to classify the ambiguous pattern of input data caused by the insufficient hidden neurons. In such cases, a new hidden neuron is added to the RBM to split the fluctuated neuron by inheritance of the parent hidden neuron attributes. In order to the insufficient hidden neurons, the condition of neuron generation is defined by monitoring the inner product of variance of monitoring both of 2 parameters $\bvec{c}$ and $\bvec{W}$ as shown in Eq.(\ref{eq:neuron_generation}).

\begin{equation}
(\alpha_{c} \cdot dc_j) \cdot (\alpha_{W} \cdot dW_{ij} )> \theta_{G},
\label{eq:neuron_generation}
\end{equation}
where $dc_j$ is the gradient of the hidden neuron $j$ and $dW_{ij}$ is the gradient of the weight vector between the neuron $i$ and the neuron $j$. In Eq.\ref{eq:neuron_generation}, there are the coefficinets $\alpha_{c}$ and $\alpha_{W}$ to regulate the range of variance of $\{c, W\}$. $\theta_{G}$ is an threshold value to make a new hidden neuron. Fig.~\ref{fig:neuron_generation} shows the neuron generation process. A new neuron in hidden layer will be generated if Eq.(\ref{eq:neuron_generation}) is satisfied and the neuron is inserted into the neighbor position of the parent neuron.

After the neuron generation process, the network with some inactivated neurons that don't contribute to the classification capability should be removed in terms of the reduction of calculation. If Eq.(\ref{eq:neuron_annihilation}) is satisfied in learning phase, the corresponding neuron will be annihilated as shown in Fig.~\ref{fig:neuron_annihilation}. 

\begin{equation}
\frac{1}{N}\sum_{n=1}^{N} p(h_j = 1 | \bvec{v}_n) < \theta_{A},
\label{eq:neuron_annihilation}
\end{equation}
where $\bvec{v}_{n}=\{ \bvec{v}_{1},\bvec{v}_{2},\cdots,\bvec{v}_{N}\}$ is a given input data where $N$ is the size of input neurons. $p(h_j = 1 | \bvec{v}_n)$ is a conditional probability of $h_j \in \{ 0, 1 \}$ under a given $\bvec{v}_n$. $\theta_{A}$ is an appropriate threshold value. The proposed method by the neuron generation algorithm and the neruron annihilation algorithm showed the good performance for some benchmark tests in previous work \cite{Kamada16a}.

\subsection{Structural Learning Method with Forgetting}
Although the optimal structure of hidden layer is obtained by the neuron generation and neuron annihilation algorithms, we may meet another difficulty that the trained network is still a black box. In other words, the knowledge extraction from the trained network can not be realized. The explicit knowledge can throw strong light on `black box' to discuss the regularities of the RBM network structure. In the subsection, we introduced the Structural Learning Method with Forgetting (SLF) to discover the regularities of the network structure in RBM \cite{Kamada16_ICONIP}. The basic concept is derived by Ishikawa \cite{Ishikawa96}, 3 kinds of penalty terms are added into an original energy function $J$ as shown in Eq.(\ref{eq:forgetting1}) -  Eq.(\ref{eq:forgetting3}). They are `learning with forgetting', `hidden units clarification,' and `learning with selective forgetting', respectively. Eq.(\ref{eq:forgetting1}) decays the weights norm multiplied by small criterion $\epsilon_{1}$. Eq.(\ref{eq:forgetting2}) forces each hidden unit to be in $\{0, 1\}$. After Eq.(\ref{eq:forgetting2}), the characteristic patterns can be extracted from the network structure. Eq.(\ref{eq:forgetting3}) works to prevent the larger energy function.
\begin{equation}
  \label{eq:forgetting1}
  J_{f} = J + \epsilon_{1} \| \bvec{W} \|,
\end{equation}
\begin{equation}
  \label{eq:forgetting2}
  J_{h} = J + \epsilon_{2} \sum_{i} \min \{ 1 - h_i, h_i\},
\end{equation}
\vspace{-2mm}
\begin{eqnarray}
  \label{eq:forgetting3}
  J_{s} = J - \epsilon_{3} \| \bvec{W}^{'} \|, \ 
  W^{'}_{ij} = \left\{
  \begin{array}{ll}
    W_{ij}, & if \ |W_{ij}| < \theta \\
    0, & otherwise
  \end{array}
  \right.,
\end{eqnarray}
where $\epsilon_{1}$, $\epsilon_{2}$ and $\epsilon_{3}$ are the small criterion at each equation. After the optimal structure of hidden layer is determined by neuron generation and neuron annihilation process, both Eq.(\ref{eq:forgetting1}) and Eq.(\ref{eq:forgetting2}) are applied together during a certain learning period. Alternatively, Eq.(\ref{eq:forgetting3}) is used instead of Eq.(\ref{eq:forgetting1}) at final learning period so as to be the large objective function.

\section{Adaptive Learning method of DBN}
\label{sec:DBN}
Deep Belief Network (DBN) is well known deep learning method proposed by Hinton \cite{Hinton06}. The model has the hierarchical network structure where each layer is pre-trained by RBM. The hierarchical DBN network structure becomes to represent higher and multiple level features of input patterns by building up the pre-trained RBM.

In this paper, we propose the adaptive learning method of DBN that can determine the optimal structure of hidden layers. A RBM employs the adaptive learning method by the neuron generation and neuron annihilation process as described in section \ref{subsec:AdaptiveRBM}. In general, data representation of DBN performs the specified features from abstract to concrete at each layer in the direction to output layer. That is, the lower layer has the power of non figurative representation, and the higher layer constructs the object to figure out an image of input patterns. Adaptive DBN can automatically adjust self-organization of structured data representation.

In the learning process of adaptive DBN, we observe the total WD (the variance of both $\bvec{c}$ and $\bvec{W}$) and energy function. If the overall WD is larger than a threshold value and the energy function is still large, then a new RBM is required to express the suitable network structure for the given input data. That is, the large WD and the large energy function indicate the condition that the RBM has lack data representation capability to figure out an image of input patterns in the overall network. Therefore, we define the condition of layer generation with the total WD and the energy function as shown in Eq.(\ref{eq:layer_generation1}) and Eq.(\ref{eq:layer_generation2}).
\begin{equation}
\sum_{l=1}^{k} (\alpha_{WD} \cdot WD^{l}) > \theta_{L1},
\label{eq:layer_generation1}
\end{equation}
\vspace{-2mm}
\begin{equation}
\sum_{l=1}^{k} (\alpha_{E} \cdot E^{l}) > \theta_{L2},
\label{eq:layer_generation2}
\end{equation}
where $WD^{l}$ is the total variance of parameters $\bvec{c}$ and $\bvec{W}$ in $l$-th RBM. $E^{l}$ is the total energy function in $l$-th RBM. $k$ means the current layer. $\alpha_{WD}$ and $\alpha_{E}$ are the constants for the adjustment of deviant range in $WD^{l}$ and $E^{l}$. $\theta_{L1}$ and $\theta_{L2}$ are the appropriate threshold values. If Eq.(\ref{eq:layer_generation1}) and Eq.(\ref{eq:layer_generation2}) are satisfied simultaneously in learning phase at layer $k$, a new hidden layer $k+1$ will be generated after the learning at layer $k$ is finished. The initial RBM parameters $\bvec{b}$, $\bvec{c}$ and $\bvec{W}$ at the generated layer $k+1$ are set by inheriting the attributes of the parent(lower) RBM.

\section{Experimental Results}
\label{sec:EXE}
\subsection{Data Sets}
The image benchmark data set, ``CIFAR-10'' and ``CIFAR-100'' \cite{CIFAR10}, was used in this paper. CIFAR-10 is about 60,000 color images data set included in about 50,000 training cases with 10 classes and about 10,000 test cases with 100 classes. The number of images in CIFAR-100 is also same as CIFAR-10, but CIFAR-100 has much more categories than CIFAR-10. The original image data with ZCA whitening are used in \cite{Dieleman12}.

We explain the parameters of RBM in this experiments. Stochastic Gradient Descent (SGD) method, 100 batch size, 0.1 constant learning rate was used. The RBM with 300 hidden neurons starts to train. The computation results were compared with the classification accuracy of the output layer which is added after DBN trains as Hinton introduced \cite{Hinton12}.

\subsection{Experimental Results}
Table~\ref{tab:result-correct-ratio-cifar10} and Table~\ref{tab:result-correct-ratio-cifar100} show the classification accuracy on CIFAR-10 and CIFAR-100, respectively. Our proposed DBN model showed higher classification capability than not only adaptive RBM with Forgetting but also traditional DBN, CNN model, and the results in \cite{record}. Table~\ref{tab:result-adaptiveDBN} shows the situation of each layer after training of our proposed DBN method with $\theta_G = 0.05$. As shown in Table~\ref{tab:result-adaptiveDBN}, the total energy and the total error were decreased. Moreover, the accuracy for classification of test data set was increased as a RBM is growing. Finally, 5 RBMs were automatically constructed according to the input data by the layer generation condition.

In particularly interesting, the number of hidden neurons was smaller in the odd layers and was larger in the even layers repeatedly. Namely, the first RBM classified the features of input data roughly, after a new RBM was built, the network tried to express more complex features as the combination of new classification patterns in the previous layer was formed.

\begin{figure}[tbp]
  \begin{center}
    \subfigure[Weights in odd (3) layer]{\includegraphics[scale=0.2]{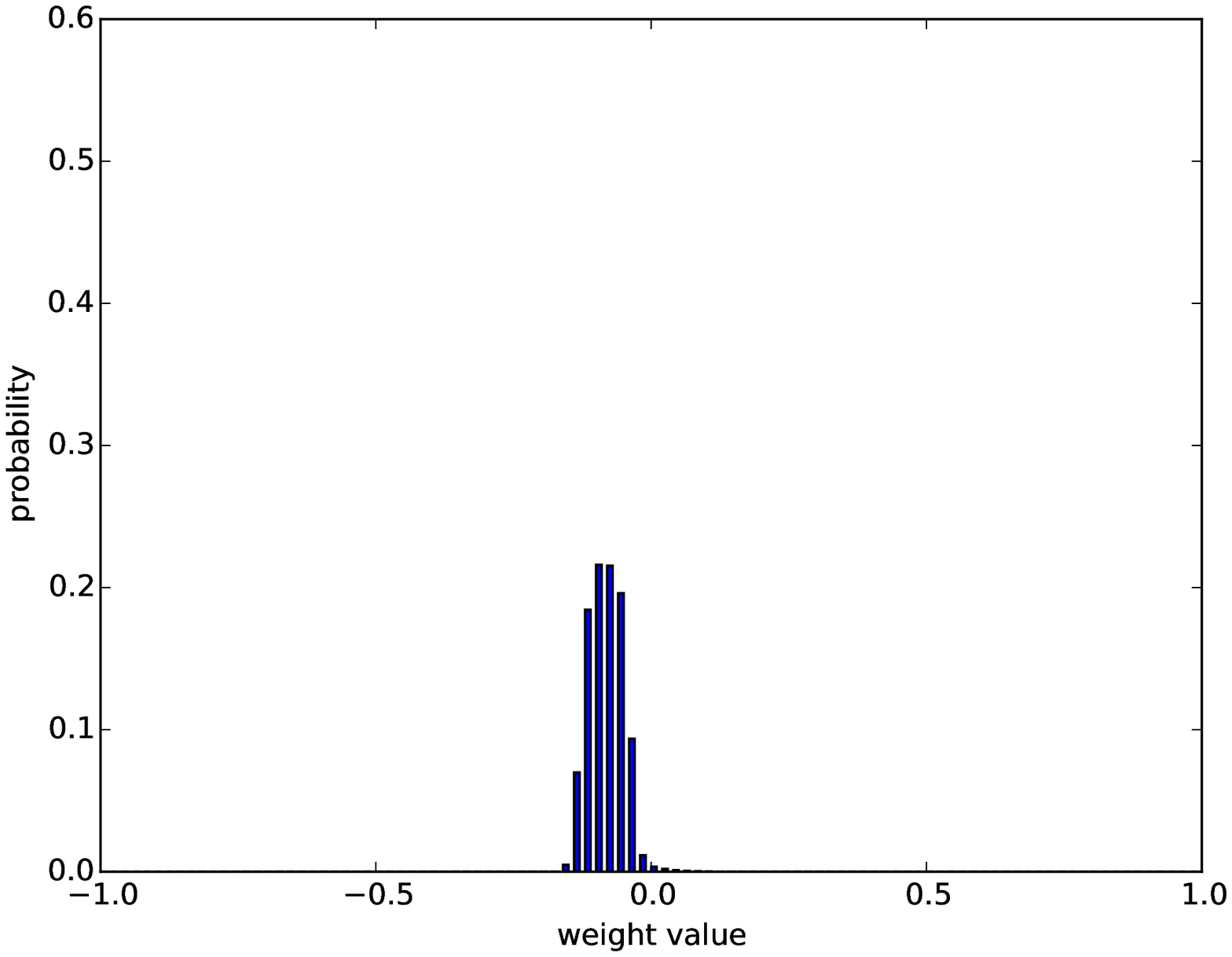}\label{fig:hist-odd-w}}
    \subfigure[Weights in even (4) layer]{\includegraphics[scale=0.2]{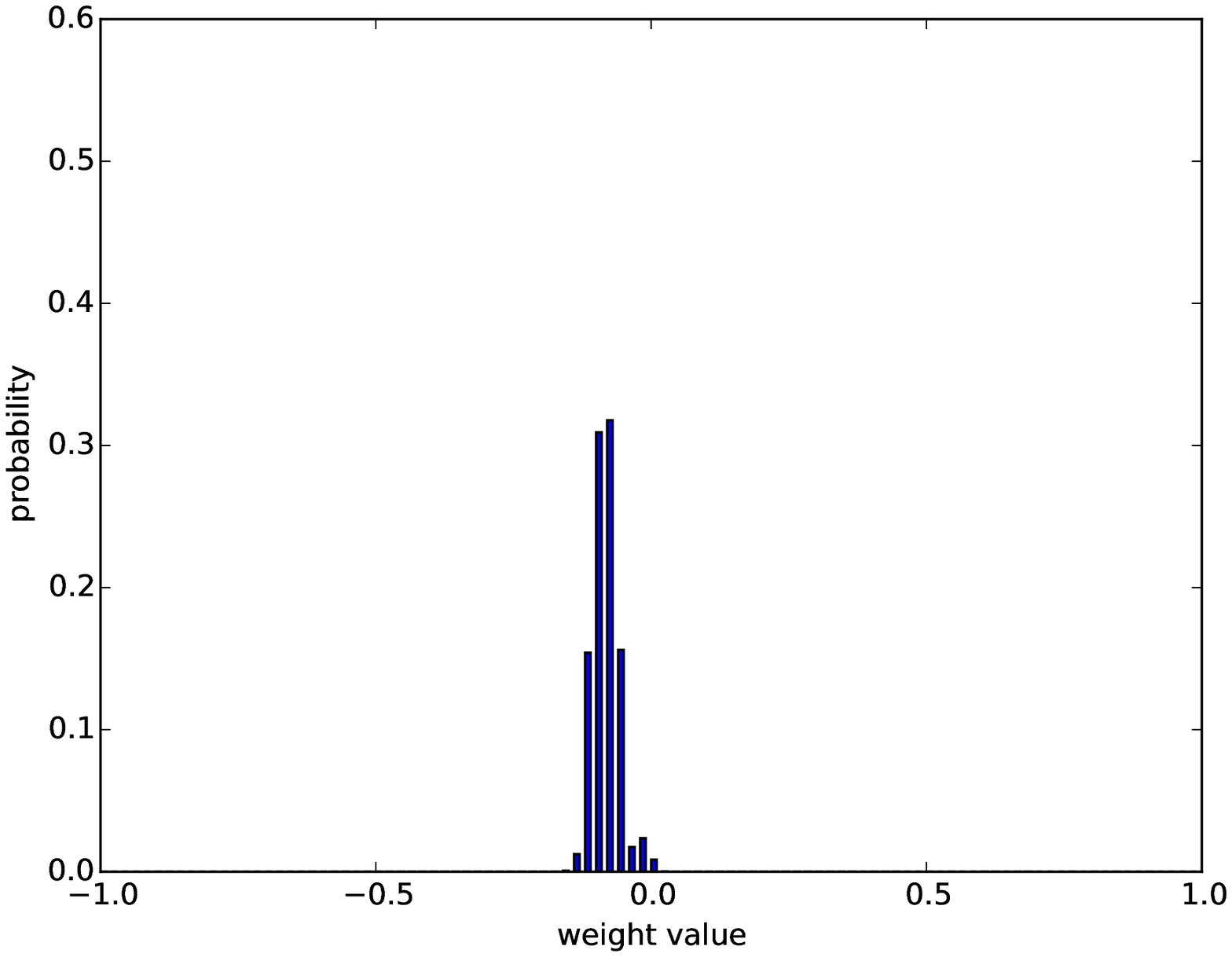}\label{fig:hist-even-w}}
    \subfigure[Outputs in odd (3) layer ]{\includegraphics[scale=0.2]{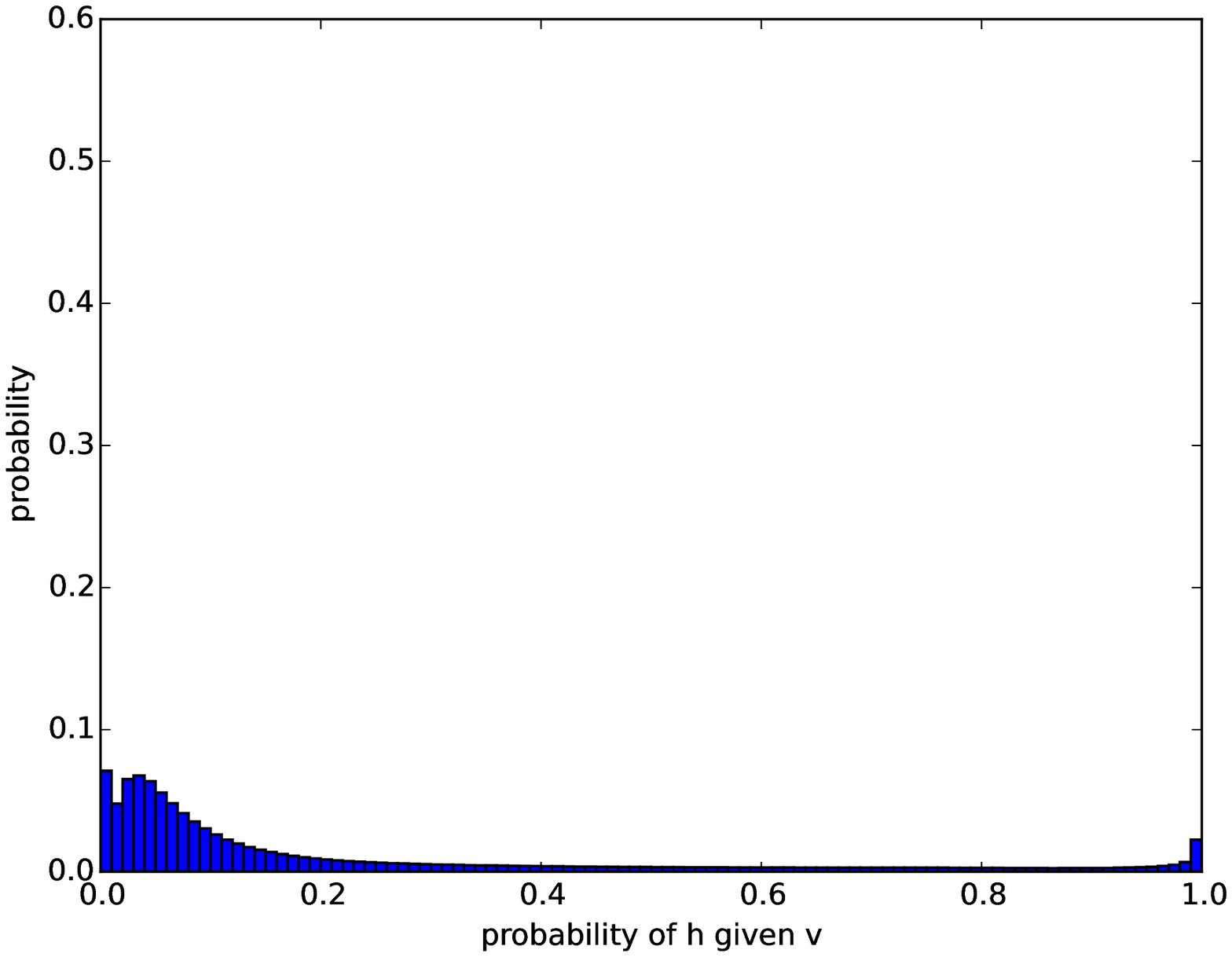}\label{fig:hist-odd-h}}
    \subfigure[Outputs in even (4) layer]{\includegraphics[scale=0.2]{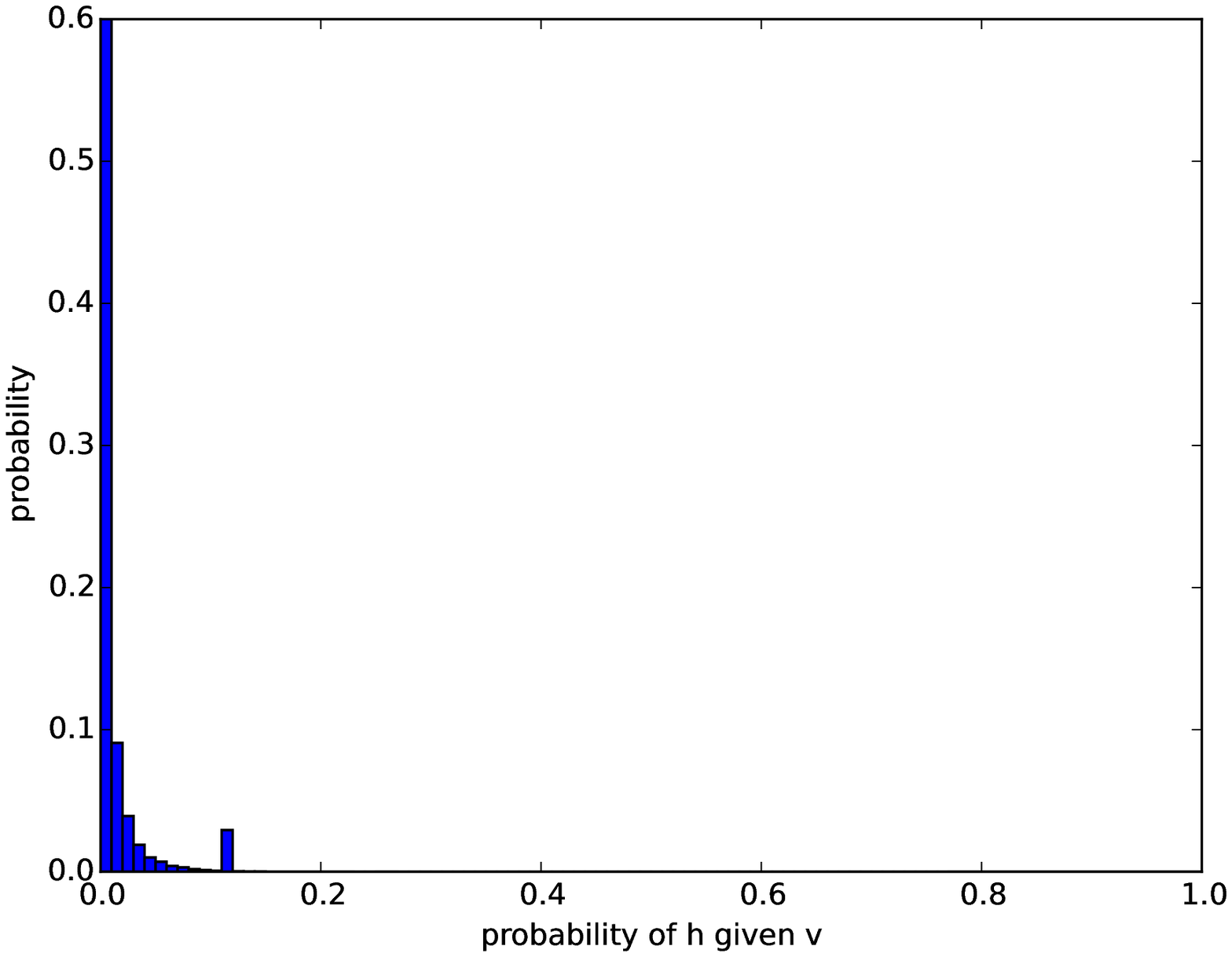}\label{fig:hist-even-h}}
  \end{center}
\caption{Weight Sizes and Outputs at Hidden Neurons}
\label{fig:result-hist}
\vspace{-5mm}
\end{figure}

Fig.\ref{fig:result-hist} shows the size of the trained weight value and the size of outputs from hidden neurons in our proposed DBN method with $\theta_G = 0.05$. We found characteristical features that outputs from hidden neurons in the even layer (with larger hidden neurons) made the sparse network structure with more districted hidden neurons as shown in Fig.\ref{fig:hist-odd-h}. On the other hand, the odd layer (with smaller hidden neurons) formed the distribution with the value except 0 or 1.

\section{Conclusion}
\label{sec:Conclusion}
The adaptive RBM method can improve the problem of the network structure according to the given input data during the learning phase. In addition, we developed the adaptive method of RBM structural learning with Forgetting to obtain a sparse structure of RBM. In this paper, we propose the layer generation condition in DBN by measuring the variance of parameters and the energy function of overall the network during the learning. In the experimental results, our proposed DBN model can reach the highest classification capability in comparison to not only the previous adaptive RBM but also traditional DBN or CNN model. Furthermore, the following interesting feature was observed. The hierarchical network structure was self-organized according to the feature of the input data as repeating of the rough classification and the detailed classification alternately upward layer and downward layer to express more complex features. The extraction method of explicit knowledge from the trained hierarchical network was considered in \cite{Kamada16_YRW}. The embedding method of the acquired knowledge into the trained network will be improved to realize approximately 100.0\% classification capability in future \cite{Kamada16_IWCIA}.

\begin{table}[h]
\caption{Classification Rate on CIFAR-10}
\vspace{-5mm}
\label{tab:result-correct-ratio-cifar10}
\begin{center}
\begin{tabular}{l|r|r}
\hline \hline
& \multicolumn{1}{c|}{Training} & \multicolumn{1}{c}{Test}  \\ \hline\hline
Traditional RBM \cite{Dieleman12} &     -     & 63.0\%  \\ \hline
Traditional DBN \cite{Krizhevsky10}      &     -     & 78.9\%  \\ \hline 
CNN \cite{Clevert15}      &     -     & 96.53\%  \\ \hline \hline
Adaptive RBM                 &   99.9\%  & 81.2\%   \\ \hline
Adaptive RBM with Forgetting  &   99.9\%  & 85.8\%   \\ \hline
Adaptive DBN with Forgetting &  100.0\%  & 92.4\%   \\
(No. layer = 5, $\theta_G = 0.05$) & & \\\hline
Adaptive DBN with Forgetting&  100.0\%  & {\bf 97.1\%}   \\
(No. layer = 5, $\theta_G = 0.01$ ) & & \\\hline
\hline 
\end{tabular}
\end{center}

\vspace{-8mm}
\end{table}
\begin{table}[h]
\caption{Classification Rate on CIFAR-100}
\vspace{-5mm}
\label{tab:result-correct-ratio-cifar100}
\begin{center}
\begin{tabular}{l|r|r}
\hline \hline
& \multicolumn{1}{c|}{Training} & \multicolumn{1}{c}{Test}  \\ \hline\hline
CNN \cite{Clevert15}     &     -       & 75.7\%  \\ \hline \hline
Adaptive DBN with Forgetting &  100.0\%  & 78.2\%   \\
(No. layer = 5, $\theta_G = 0.05$) & & \\\hline
Adaptive DBN with Forgetting&  100.0\%  & {\bf 81.3\%}   \\
(No. layer = 5, $\theta_G = 0.01$ ) & & \\\hline
\hline 
\end{tabular}
\end{center}
\end{table}

\vspace{-8mm}
\begin{table}[htb]
  \begin{center}
    \caption{Situation in each layer after training of adaptive DBN}
    \label{tab:result-adaptiveDBN}
\vspace{-3mm}
    \begin{tabular}{l|r|r|r|r}\hline\hline
      Layer & No. neurons & Total energy & Total error & Accuracy\\ \hline \hline
      1 & 433 & -0.24  & 25.37  & 84.6\% \\ \hline
      2 & 1595 & -1.01 & 10.76  & 86.2\% \\ \hline
      3 & 369 & -0.78  &  1.77 & 90.6\% \\ \hline        
      4 & 1462 & -1.00 &  0.43 & 92.3\% \\ \hline
      5 & 192 & -1.17  &  0.01 & {\bf 92.4\%} \\ \hline
      \hline
    \end{tabular}
  \end{center}
\end{table}
\vspace{-5mm}

\section*{Acknowledgment}
This research and development work was supported by the MIC/SCOPE \#162308002.


\begin{thebibliography}{1}

\bibitem{Bengio09}
Y.Bengio, \emph{Learning Deep Architectures for AI}. Foundations and Trends in Machine Learning archive, Vol.2, No.1, pp.1--127 (2009)

\bibitem{Hinton12}
G.E.Hinton, \emph{A Practical Guide to Training Restricted Boltzmann Machines}. Neural Networks, Tricks of the Trade, Lecture Notes in Computer Science, Vol.7700, pp.599--619 (2012)

\bibitem{Hinton06}
G.E.Hinton, S.Osindero and Y.Teh, \emph{A fast learning algorithm for deep belief nets}. Neural Computation, Vol.18, No.7, pp.1527--1554 (2006)

\bibitem{Kamada16_ICONIP}
S.Kamada, T.Ichimura,
`A Structural Learning Method of Restricted Boltzmann Machine by Neuron Generation and Annihilation Algorithm',
\emph{Proc. of the 23rd International Conference on Neural Information Processing (Springer LNCS9950)}, pp.372-380 (2016)

\bibitem{CIFAR10}
A.Krizhevsky,
`Learning Multiple Layers of Features from Tiny Images',
\emph{Master of thesis, University of Toronto} (2009)

\bibitem{Kamada16_YRW}
S.Kamada, T.Ichimura,
`A Consideration of Knowledge Acquisition from Adaptive Learning Method of Deep Belief Network',
\emph{Proc. 2016 IEEE Young Researcher's Workshop},
IEEE SMC Hiroshima Chapter, pp.61-66 (Japanese) (2016)

\bibitem{record}
Classification datasets results, 
\url{http://rodrigob.github.io/are_we_there_yet/build/classification_datasets_results.html}, [online] (2016)

\bibitem{Hinton02}
G.E.Hinton, \emph{Training products of experts by minimizing contrastive divergence}. Neural Computation, Vol.14, pp.1771--1800 (2002)

\bibitem{Carlson15}
D.Carlson, V.Cevher and L.Carin, \emph{Stochastic Spectral Descent for Restricted Boltzmann Machines}. Proc. of the Eighteenth International Conference on Artificial Intelligence and Statistics, pp.111-119 (2015)

\bibitem{Kamada16a}
S.Kamada and T.Ichimura, \emph{An Adaptive Learning Method of Restricted Boltzmann Machine by Neuron Generation and Annihilation Algorithm}. Proc. of 2016 IEEE SMC (SMC 2016), (to appear in 2016)

\bibitem{Ichimura04}
T.Ichimura and K.Yoshida Eds., \emph{Knowledge-Based Intelligent Systems for Health Care}. Advanced Knowledge International (ISBN 0-9751004-4-0) (2004)

\bibitem{Ishikawa96}
M.Ishikawa, \emph{Structural Learning with Forgetting}. Neural Networks, Vol.9, No.3, pp.509-521 (1996)

\bibitem{Dieleman12}
S.Dieleman and B.Schrauwen, \emph{Accelerating sparse restricted Boltzmann machine training using non-Gaussianity measures}. Deep Learning and Unsupervised Feature Learning (NIPS-2012) (2012)

\bibitem{Krizhevsky10}
A.Krizhevsky, \emph{A Convolutional Deep Belief Networks on CIFAR-10}, Technical report (2010)

\bibitem{Clevert15}
D.A.Clevert, T.Unterthiner, and S.Hochreiter, 
\emph{Fast and Accurate Deep Network Learning by Exponential Linear Units (ELUs)},
Proc. of ICRL2016 (2016)

\bibitem{Kamada16_IWCIA}
S.Kamada, T.Ichimura,
\emph{Fine Tuning Method by using Knowledge Acquisition from Deep Belief Network},
Proc. of IEEE 8th International Workshop on Computational Intelligence and Applications (IWCIA2016) (to appear in 2016)
  
\end{thebibliography}
\end{document}